\documentclass[runningheads]{llncs}
\usepackage{graphicx}
\usepackage{comment}
\usepackage{amsmath,amssymb} % define this before the line numbering.
\usepackage{color}
\usepackage[dvipsnames]{xcolor}
\usepackage{tablefootnote}

\makeatletter
\newcommand{\printfnsymbol}[1]{%
  \textsuperscript{\@fnsymbol{#1}}%
}
\makeatother

\usepackage{subcaption}
\begin{document}
% \renewcommand\thelinenumber{\color[rgb]{0.2,0.5,0.8}\normalfont\sffamily\scriptsize\arabic{linenumber}\color[rgb]{0,0,0}}
% \renewcommand\makeLineNumber {\hss\thelinenumber\ \hspace{6mm} \rlap{\hskip\textwidth\ \hspace{6.5mm}\thelinenumber}}
% \linenumbers
\pagestyle{headings}
\mainmatter
\def\ECCVSubNumber{3992}  % Insert your submission number here

\title{StyleGAN2 Distillation for Feed-forward Image~Manipulation} % Replace with your title

% INITIAL SUBMISSION 
\begin{comment}
% \titlerunning{ECCV-20 submission ID \ECCVSubNumber} 
% \authorrunning{ECCV-20 submission ID \ECCVSubNumber} 
% \author{Anonymous ECCV submission}
% \institute{Paper ID \ECCVSubNumber}
\end{comment}
%******************

% CAMERA READY SUBMISSION
% \begin{comment}
\titlerunning{StyleGAN2 Distillation for Feed-forward Image Manipulation}
% If the paper title is too long for the running head, you can set
% an abbreviated paper title here

%
\author{Yuri Viazovetskyi\thanks{equal contribution}\inst{1} \and
Vladimir Ivashkin\printfnsymbol{1}\inst{1,2} \and
Evgeny Kashin\printfnsymbol{1}\inst{1}}
\authorrunning{Y. Viazovetskyi et al.}
% First names are abbreviated in the running head.
% If there are more than two authors, 'et al.' is used.
%
\institute{
    Yandex \and
    Moscow Institute of Physics and Technology \\
    \email{\{iviazovetskyi,vlivashkin,evgenykashin\}@yandex-team.ru}
}
% \end{comment}
%******************
\maketitle
\begin{figure}
    \centering
    \includegraphics[width=.99\linewidth]{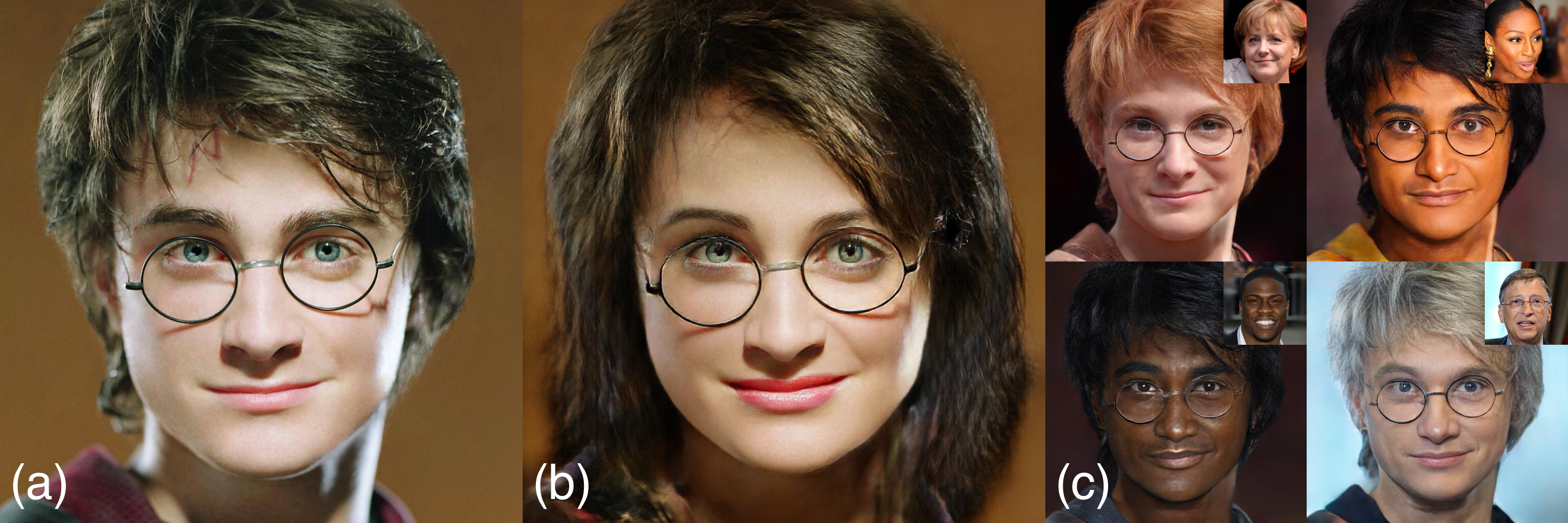}
    \caption{Image manipulation examples generated by our method from (a) source image sampled from Celeba-HQ: (b) gender swap at 1024x1024 and (c) style mixing at 512x512. Samples are generated feed-forward, StyleGANv2 which we distilled was trained on FFHQ}
    \label{fig:title}
\end{figure}
\begin{abstract}

StyleGAN2 is a state-of-the-art network in generating realistic images. Besides, it was explicitly trained to have disentangled directions in latent space, which allows efficient image manipulation by varying latent factors. Editing existing images requires embedding a given image into the latent space of StyleGAN2. Latent code optimization via backpropagation is commonly used for qualitative embedding of real world images, although it is prohibitively slow for many applications. We propose a way to distill a particular image manipulation of StyleGAN2 into image-to-image network trained in paired way. The resulting pipeline is an alternative to existing GANs, trained on unpaired data. We provide results of human faces’ transformation: gender swap, aging/rejuvenation, style transfer and image morphing. We show that the quality of generation using our method is comparable to StyleGAN2 backpropagation and current state-of-the-art methods in these particular tasks.
\keywords{Computer Vision, StyleGAN2, distillation, synthetic data}
% TLDR: Paired image-to-image translation, trained on synthetic data generated by StyleGAN2 outperforms existing approaches in image manipulation
\end{abstract}

\section{Introduction}
Generative adversarial networks (GANs) \cite{goodfellow2014generative} have created wide opportunities in image manipulation.
General public is familiar with them from the many applications which offer to change one's face in some way: make it older/younger, add glasses, beard, etc.

There are two types of network architecture which can perform such translations feed-forward: neural networks trained on either paired or unpaired datasets. In practice, only unpaired datasets are used. The methods used there are based on cycle consistency \cite{zhu2017unpaired}. The follow-up studies  \cite{huang2018multimodal,choi2018stargan,choi2019stargan} have maximum resolution of 256x256.

At the same time, existing paired methods (e.g. pix2pixHD \cite{wang2018high} or SPADE \cite{park2019semantic}) support resolution up to 2048x1024. But it is very difficult or even impossible to collect a paired dataset for such tasks as age manipulation. For each person, such dataset would have to contain photos made at different age, with the same head position and facial expression. Close examples of such datasets exist, e.g. CACD \cite{chen2014cross}, AgeDB \cite{moschoglou2017agedb}, although with different expressions and face orientation. To the best of our knowledge, they have never been used to train neural networks in a paired mode.

These obstacles can be overcome by making a synthetic paired dataset, if we solve two known issues concerning dataset generation: appearance gap \cite{hoffman2017cycada} and content gap \cite{kar2019meta}. 
Here, unconditional generation methods, like StyleGAN \cite{karras2019style}, can be of use. StyleGAN generates images of quality close to real world and with distribution close to real one according to low FID results. Thus output of this generative model can be a good substitute for real world images. The properties of its latent space allow to create sets of images differing in particular parameters.
Addition of path length regularization (introduced as measure of quality in \cite{karras2019style}) in the second version of StyleGAN \cite{karras2019analyzing} makes latent space even more suitable for manipulations. 

Basic operations in the latent space correspond to particular image manipulation operations. Adding a vector, linear interpolation, and crossover in latent space lead to expression transfer, morphing, and style transfer, respectively.
The distinctive feature of both versions of StyleGAN architecture is that the latent code is applied several times at different layers of the network. Changing the vector for some layers will lead to changes at different scales of generated image. Authors group spatial resolutions in process of generation into coarse, middle, and fine ones. It is possible to combine two people by using one person's code at one scale and the other person's at another.

Operations mentioned above are easily performed for images with known embeddings. For many entertainment purposes this is vital to manipulate some existing real world image on the fly, e.g. to edit a photo which has just been taken. Unfortunately, in all the cases of successful search in latent space described in literature the backpropagation method was used~\cite{abdal2019image2stylegan++,abdal2019image2stylegan,gabbay2019style,karras2019analyzing,shen2020interpreting}. Feed-forward is only reported to be working as an initial state for latent code optimization \cite{baylies2019styleganencoder}. Slow inference makes application of image manipulation with StyleGAN2 in production very limited: it costs a lot in data center and is almost impossible to run on a device. However, there are examples of backpropagation run in production, e.g. \cite{shi2019face}.

In this paper we consider opportunities to distill \cite{hinton2015distilling,ba2014deep} a particular image manipulation of StyleGAN2 generator, trained on the FFHQ dataset. The distillation allows to extract the information about faces' appearance and the ways they can change (e.g. aging, gender swap) from StyleGAN into image-to-image network. We propose a way to generate a paired dataset and then train a ``student'' network on the gathered data. This method is very flexible and is not limited to the particular image-to-image model.

Despite the resulting image-to-image network is trained only on generated samples, we show that it performs on real world images on par with StyleGAN backpropagation and current state-of-the-art algorithms trained on unpaired data.

Our contributions are summarized as follows:
\begin{itemize}
    \item We create synthetic datasets of paired images to solve several tasks of image manipulation on human faces: gender swap, aging/rejuvenation, style transfer and face morphing;
    \item We show that it is possible to train image-to-image network on synthetic data and then apply it to real world images;
    \item We study the qualitative and quantitative performance of image-to-image networks trained on the synthetic datasets;
    \item We show that our approach outperforms existing approaches in gender swap task.
\end{itemize}

We publish all collected paired datasets for reproducibility and future research: \url{https://github.com/EvgenyKashin/stylegan2-distillation}.

\section{Related work}
\subsubsection{Unconditional image generation}
Following the success of ProgressiveGAN \cite{karras2017progressive} and BigGAN \cite{brock2018large}, StyleGAN \cite{karras2019style} became state-of-the-art image generation model. This was achieved due to rethinking generator architecture and borrowing approaches from style transfer networks: mapping network and AdaIN~\cite{huang2017arbitrary}, constant input, noise addition, and mixing regularization. The next version of StyleGAN -- StyleGAN2 \cite{karras2019analyzing}, gets rid of artifacts of the first version by revising AdaIN and improves disentanglement by using perceptual path length as regularizer. 

Mapping network is a key component of StyleGAN, which allows to transform latent space $\mathcal{Z}$ into less entangled intermediate latent space $\mathcal{W}$. Instead of actual latent $z \in \mathcal{Z}$ sampled from normal distribution, $w \in \mathcal{W}$ resulting from mapping network $f: \mathcal{Z} \rightarrow \mathcal{W}$ is fed to AdaIN.
Also it is possible to sample vectors from extended space $\mathcal{W}+$, which consists of multiple independent samples of $\mathcal{W}$, one for each layer of generator. Varying $w$ at different layers will change details of generated picture at different scales.

\subsubsection{Latent codes manipulation}
It was recently shown \cite{Goetschalckx_2019_ICCV,jahanian2019steerability} that linear operations in latent space of generator allow successful image manipulations in a variety of domains and with various GAN architectures. 
In GANalyze \cite{Goetschalckx_2019_ICCV}, the attention is directed to search interpretable directions in latent space of BigGAN~\cite{brock2018large} using MemNet~\cite{khosla2015understanding} as ``assessor'' network.
Jahanian et al. \cite{jahanian2019steerability} show that walk in latent space lead to interpretable changes in different model architectures: BigGAN, StyleGAN, and DCGAN~\cite{radford2015unsupervised}.

To manipulate real images in latent space of StyleGAN, one needs to find their embeddings in it. The method of searching the embedding in intermediate latent space via backprop optimization is described in \cite{abdal2019image2stylegan++,abdal2019image2stylegan,gabbay2019style,shen2020interpreting}. The authors use non-trivial loss functions to find both close and perceptually good image and show that embedding fits better in extended space $\mathcal{W}+$. Gabbay et al.~\cite{gabbay2019style} show that StyleGAN generator can be used as general purpose image prior. Shen et al.~\cite{shen2020interpreting} show the opportunity to manipulate appearance of generated person, including age, gender, eyeglasses, and pose, for both PGGAN~\cite{karras2017progressive} and StyleGAN. The authors of StyleGAN2~\cite{karras2019analyzing} propose to search embeddings in $\mathcal{W}$ instead of $\mathcal{W}+$ to check if the picture was generated by StyleGAN2.

\subsubsection{Paired Image-to-image translation}
Pix2pix \cite{isola2017image} is one of the first conditional generative models applied for image-to-image translation. It learns mapping from input to output images.
Chen and Koltun~\cite{chen2017photographic} propose the first model which can synthesize 2048x1024 images. It is followed by pix2pixHD~\cite{wang2018high} and SPADE~\cite{park2019semantic}.
In SPADE generator, each normalization layer uses the segmentation mask to modulate the layer activations. So its usage is limited to the translation from segmentation maps. There are numerous follow-up works based on pix2pixHD architecture, including those working with video \cite{chan2019everybody,wang2019fewshotvid2vid,wang2018video}.

\subsubsection{Unpaired Image-to-image translation}
The idea of applying cycle consistency to train on unpaired data is first introduced in CycleGAN~\cite{zhu2017unpaired}. The methods of unpaired image-to-image translation can be either single mode GANs~\cite{zhu2017unpaired,Yi_2017_ICCV,liu2017unsupervised,choi2018stargan} or multimodal GANs~\cite{zhu2017toward,huang2018multimodal,lee2018diverse,DRIT_plus,liu2019few,choi2019stargan}. FUNIT~\cite{liu2019few} supports multi-domain image translation using a few reference images from a target domain. StarGAN v2~\cite{choi2019stargan} provide both latent-guided and reference-guided synthesis. All of the above-mentioned methods operate at resolution of at most 256x256 when applied to human faces.

Gender swap is one of well-known tasks of unsupervised image-to-image translation \cite{choi2018stargan,choi2019stargan,liu2019attribute}. 

Face aging/rejuvenation is a special task which gets a lot of attention \cite{zhang2017age,song2018dual,he2019s2gan}.
Formulation of the problem can vary. The simplest version of this task is making faces look older or younger~\cite{choi2018stargan}. More difficult task is to produce faces matching particular age intervals~\cite{li2018global,wang2018face,yang2019learning,liu2019attribute}. $\text{S}^2$GAN \cite{he2019s2gan} proposes continuous changing of age using weight interpolation between transforms which correspond to two closest age groups.

\subsubsection{Training on synthetic data}
Synthetic datasets are widely used to extend datasets for some analysis tasks (e.g. classification).
In many cases, simple graphical engine can be used to generate synthetic data. To perform well on real world images, this data need to overcome both appearance gap~\cite{hoffman2017cycada,french2017self,tobin2017domain,tsai2018learning,shrivastava2017learning} and content gap~\cite{kar2019meta,ruiz2018learning}.

Ravuri et al.~\cite{ravuri2019classification} study the quality of a classificator trained on synthetic data generated by BigGAN and show~\cite{ravuri2019seeing} that BigGAN does not capture the ImageNet~\cite{imagenet_cvpr09} data distributions and is only partly successful for data augmentation. Shrivastava et al.~\cite{shrivastava2017learning} reduce the quality drop of this approach by revising train setup.
Chen et al.~\cite{chen2017makeup} make paired dataset with image editing applications to train image2image network. 

Synthetic data is what underlies knowledge distillation, a technique that allows to train ``student'' network using data generated by ``teacher'' network \cite{hinton2015distilling,ba2014deep}. Usage of this additional source of data can be used to improve measures \cite{xie2019self} or to reduce size of target model \cite{mirzadeh2019improved}.
Aguinaldo et al. \cite{aguinaldo2019compressing} show that knowledge distillation is successfully applicable for generative models.

\section{Method overview}
\subsection{Data collection}

\begin{figure}
    \centering
    \includegraphics[width=.99\linewidth]{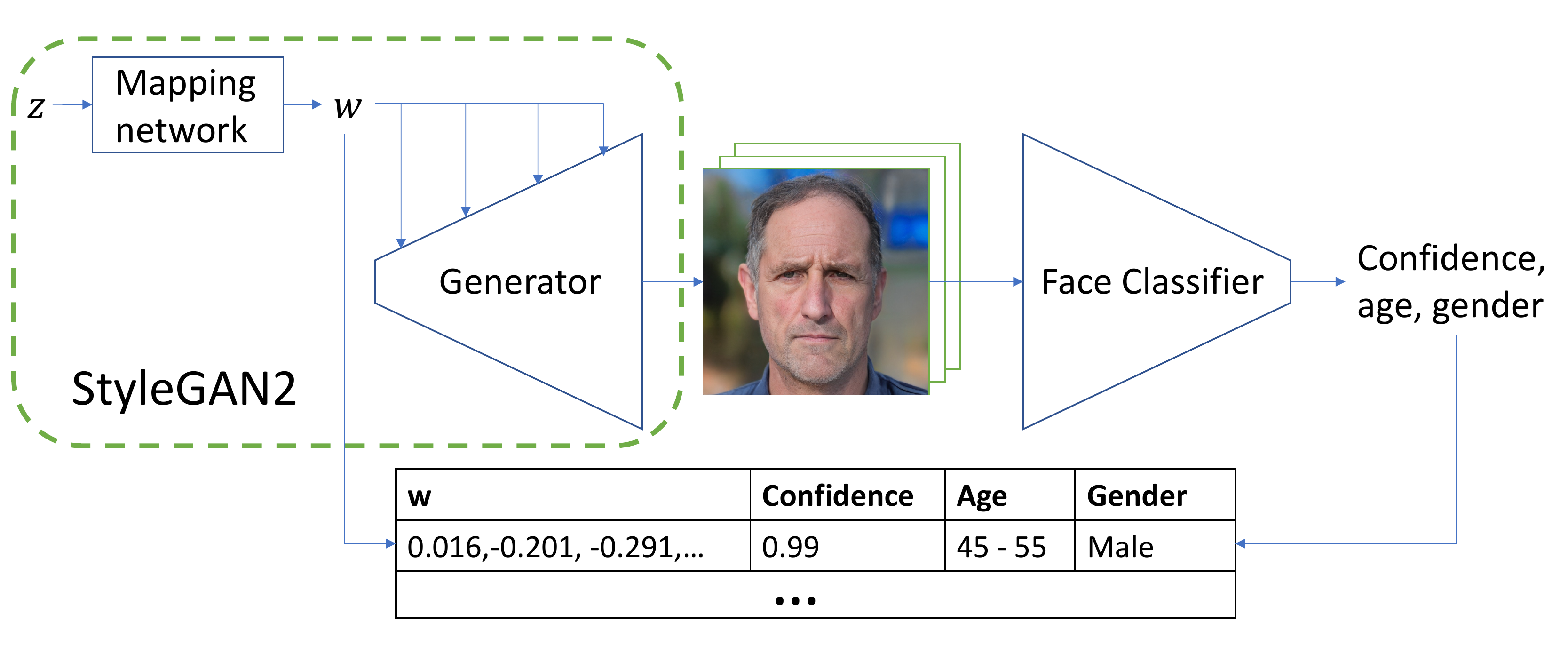}
    \caption{Method of finding correspondence between latent codes and facial attributes}
    \label{fig:method_1}
\end{figure}
All of the images used in our datasets are generated using the official implementation of StyleGAN2\footnote{\url{https://github.com/NVlabs/stylegan2}}. In addition to that we only use the config-f version checkpoint pretrained by the authors of StyleGAN2 on FFHQ dataset. All the manipulations are performed with the disentangled image codes $w$. 

We use the most straightforward way of generating datasets for style mixing and face morphing. Style mixing is described in  \cite{karras2019style} as a regularization technique and requires using two intermediate latent codes ${w_1}$ and ${w_2}$ at different scales. Face morphing corresponds to linear interpolation of intermediate latent codes $w$. We generate 50\,000 samples for each task. Each sample consists of two source images and a target image. Each source image is obtained by randomly sampling $z$ from normal distribution, mapping it to intermediate latent code ${w}$, and generating image $g(w)$ with StyleGAN2. We produce target image by performing corresponding operation on the latent codes and feeding the result to StyleGAN2.

Face attributes, such as gender or age, are not explicitly encoded in StyleGAN2 latent space or intermediate space. To overcome this limitation we use a separate pretrained face classification network. Its outputs include confidence of face detection, age bin and gender. The network is proprietary, therefore we release the final version of our gender and age datasets in order to maintain full reproducibility of this work\footnote{\url{https://github.com/EvgenyKashin/stylegan2-distillation}}. 

\begin{figure}
    \centering
    \includegraphics[width=.99\linewidth]{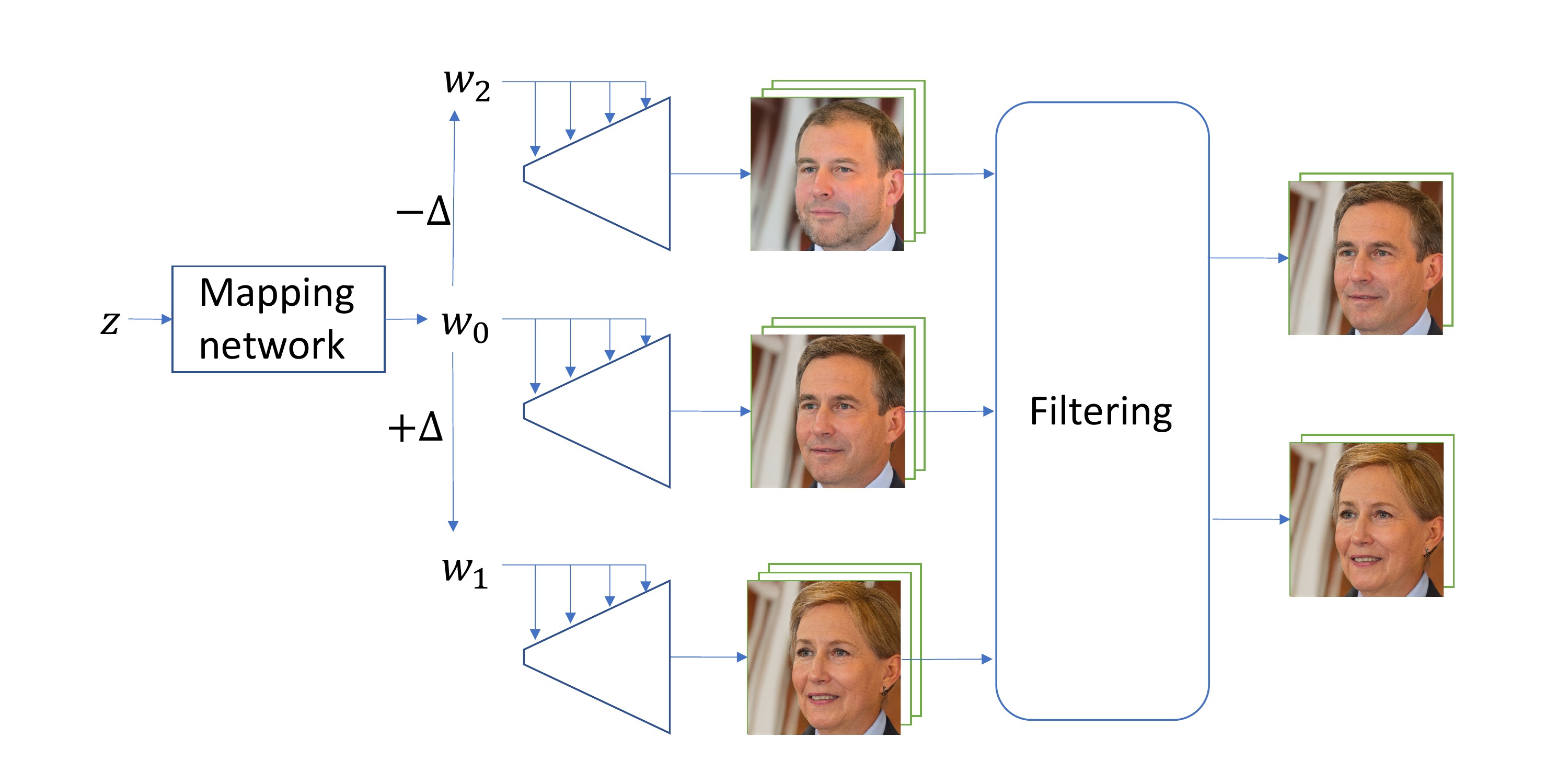}
    \caption{Dataset generation. We first sample random vectors $z$ from normal distribution. Then for each $z$ we generate a set of images along the vector $\Delta$ corresponding to a facial attribute. Then for each set of images we select the best pair based on classification results}
    \label{fig:method_2}
\end{figure}

We create gender and age datasets in four major steps. First, we generate an intermediate dataset, mapping latent vectors to target attributes as illustrated in Fig.~\ref{fig:method_1}. Second, we
find the direction in latent space associated with the attribute. Third, we generate raw dataset, using above-mentioned vector as briefly described in Fig.~\ref{fig:method_2}. Finally, we filter the images to get the final dataset. The method is described below in more detail.

\begin{enumerate}
    \item Generate random latent vectors ${z_1\dots z_n}$, map them to intermediate latent codes ${w_1\dots w_n}$, and generate corresponding image samples $g(w_i)$ with StyleGAN2.
    \item Get attribute predictions from pretrained neural network $f$, $c(w_i)=f(g(w_i))$.
    \item Filter out images where faces were detected with low confidence\footnote{This helps to reduce generation artifacts in the dataset, while maintaining high variability as opposed to lowering truncation-psi parameter.}. Then select only images with high classification certainty.
    \item Find the center of every class $C_k = \frac{1}{n_{c=k}}\sum_{c(w_i)=k} w_i$ and the transition vectors from one class to another $\Delta_{c_i, c_j} = C_j - C_i$ 
    \item Generate random samples $z_i$ and pass them through mapping network. For gender swap task, create a set of five images $g(w - \Delta), g(w - \Delta/2), g(w), g(w + \Delta/2), g(w + \Delta)$ For aging/rejuvenation first predict faces' attributes $c(w_i)$, then use corresponding vectors $\Delta_{c(w_i)}$ to generate faces that should be two bins older/younger.
    \item Get predictions for every image in the raw dataset. Filter out by confidence.
    \item From every set of images, select a pair based on classification results. Each image must belong to the corresponding class with high certainty.
\end{enumerate}

As soon as we have aligned data, a paired image-to-image translation network can be trained. 

% \todo{Examples of our generated dataset}

\subsection{Training process}
In this work, we focus on illustrating the general approach rather than solving every task as best as possible. As a result, we choose to train pix2pixHD\footnote{\url{https://github.com/NVIDIA/pix2pixHD}} \cite{wang2018high} as a unified framework for image-to-image translation instead of selecting a custom model for every type of task.

It is known that pix2pixHD has blob artifacts\footnote{\url{https://github.com/NVIDIA/pix2pixHD/issues/46}} and also tends to repeat patterns \cite{park2019semantic}. The problem with repeated patterns is solved in \cite{karras2019style,park2019semantic}. Light blobs is a problem which is solved in StyleGAN2.
We suppose that similar treatment also in use for pix2pixHD.

Fortunately, even vanilla pix2pixHD trained on our datasets produces sufficiently good results with little or no artifacts. Thus, we leave improving or replacing pix2pixHD for future work. We make most part of our experiments and comparison in 512x512 resolution, but also try 1024x1024 for gender swap.

Style mixing and face averaging tasks require two input images to be fed to the network at the same time. It is done by setting number of input channels to 6 and concatenating the inputs along channel axis.

\section{Experiments}
Although StyleGAN2 can be trained on data of different nature, we concentrate our efforts only on face data. We show application of our method to several tasks: gender swap, aging/rejuvenation and style mixing and face morphing. In all our experiments we collect data from StyleGAN2, trained on FFHQ dataset \cite{karras2019style}.

\subsection{Evaluation protocol}
Only the task of gender transform (two directions) is used for evaluation. We use Frechét inception distance (FID) \cite{heusel2017gans} for quantitative comparison of methods, as well as human evaluation.

For each feed-forward baseline we calculate FID between 50\,000 real images from FFHQ datasets and 20\,000 generated images, using 20\,000 images from FFHQ as source images. For each source image we apply transformation to the other gender, assuming source gender is determined by our classification model.
Before calculating FID measure all images are resized to 256x256 size for fair comparison.

Also human evaluation is used for more accurate comparison with optimization based methods. Our study consists of two surveys:
\begin{enumerate}
    \item \textbf{Quality.} Task for female to male translation (male to female one is similar): ``For the same image on the left, there are two different options on the right. Choose the best face, which is: turned into a male (most important), similar to the original person, the position of the face and emotions are preserved, the original items in the photo are preserved.''
    \item \textbf{Realism.} In this task, sources are different and not shown. ``Choose the image, which is: more realistic (the most important), better in quality, with fewer artifacts.''
\end{enumerate}

All images were resized to 512x512 size in this comparison.
The first task should show which method is the best at performing transformation, the second -- which looks the most real regardless of the source image. We use side-by-side experiments for both tasks where one side is our method and the other side is one of optimization based baselines. Answer choices are shuffled. For each comparison of our method with a baseline, we generate 1000 questions and each question is answered by 10 different people. For answers aggregation we use Dawid-Skene method~\cite{dawid1979maximum} and filter out the examples with confidence level less than 95\% (it is approximately 4\% of all questions).

\subsection{Distillation of image-to-image translation}
\subsubsection{Gender swap} We generate a paired dataset for male and female faces according to the method described above and than train a separate pix2pixHD model for each gender translation.

\begin{figure}
    \begin{subfigure}{.5\textwidth}
      \centering
      \includegraphics[width=.99\linewidth]{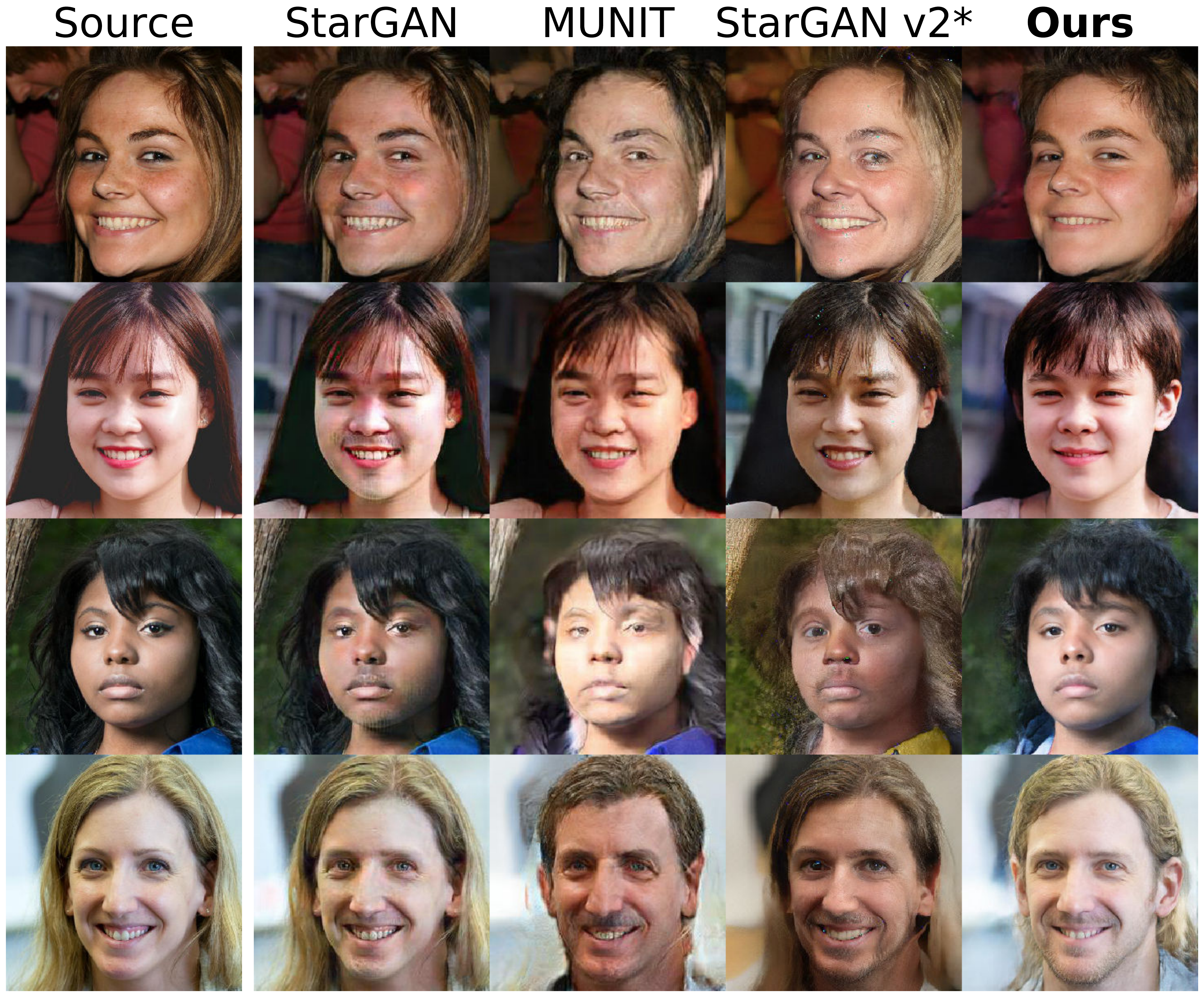}
      \caption{Female to male}
    \end{subfigure}%
    \begin{subfigure}{.5\textwidth}
      \centering
      \includegraphics[width=.99\linewidth]{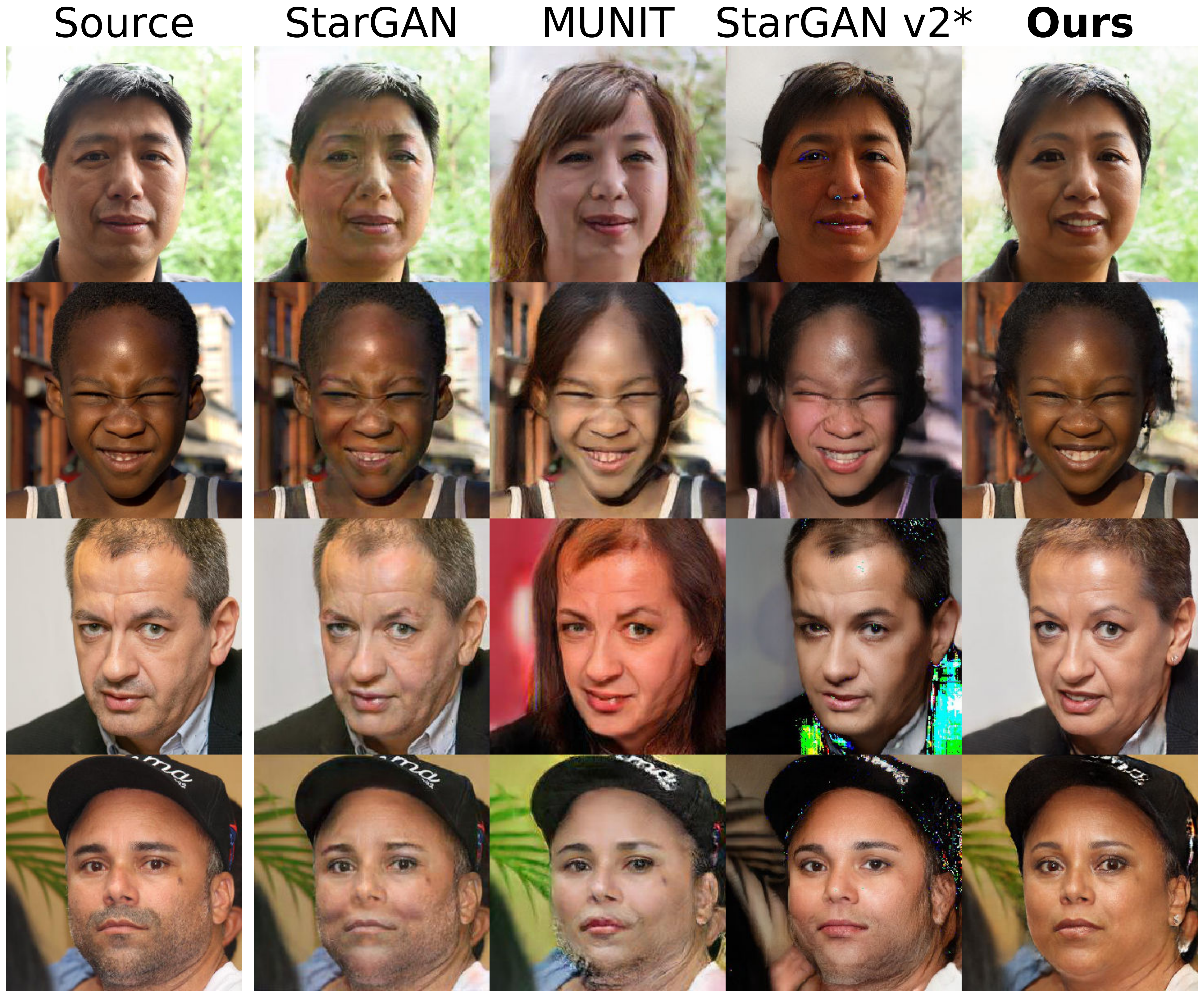}
      \caption{Male to female}
    \end{subfigure}%
    \caption{\label{fig:gender_swap_i2i} Gender transformation: comparison with image-to-image translation approaches. MUNIT and StarGAN v2* are multimodal so we show one random realization there}
\end{figure}

We compete with both unpaired image-to-image methods and different StyleGAN embedders with latent code optimization. We choose
StarGAN\footnote{\url{https://github.com/yunjey/stargan}}~\cite{choi2018stargan},
MUNIT\footnote{\url{https://github.com/NVlabs/MUNIT}} \cite{huang2018munit} and
StarGAN v2*\footnote{\url{https://github.com/taki0112/StarGAN\_v2-Tensorflow} (unofficial implementation, so its results may differ from the official one)}~\cite{choi2019stargan}
for a competition with unpaired methods. We train all these methods on FFHQ classified into males and females.

Fig.~\ref{fig:gender_swap_i2i} shows qualitative comparison between our approach and unpaired image-to-image ones. It demonstrates that distilled transformation have significantly better visual quality and more stable results. %Also, our model supports bigger resolution.
Quantitative comparison in Table~\ref{table:benchmarks-gender-ffhq} confirms our observations. We also checked that our model is perform well on other datasets without retraining. Table~\ref{table:benchmarks-gender-celeba} shows comparison of gender swap of CelebA-HQ images with models trained on CelebA. Our model wins despite it has no CelebA samples during training. The results indicate that the method can potentially be applied to real world images without retraining.

StyleGAN2 provides an official projection method. This method operates in $\mathcal{W}$, which only allows to find faces generated by this model, but not real world images. So, we also build a similar method for $\mathcal{W}+$ for comparison. It optimizes separate $w$ for each layer of the generator, which helps to better reconstruct a given image. After finding $w$ we can add transformation vector described above and generate a transformed image.

Also we add projection methods made by Dmitry Nikitko (Puzer)~\cite{puzer2019styleganencoder} and Peter Baylies (pbaylies)~\cite{baylies2019styleganencoder} for finding latent code to comparison, even though they are based on the first version of StyleGAN. These encoders are the most known implementations, they use custom perceptual losses for better perception. StyleGAN encoder by Peter Baylies is mode advanced one. In addition to more precisely selected loss functions, it uses background masking and forward pass approximation of optimization starting point.

\begin{table}
    \begin{subfigure}[t]{.5\textwidth}
      \begin{center}
        \caption{Evaluate on FFHQ}
        \label{table:benchmarks-gender-ffhq}
        \begin{tabular}{ll}
          \hline\noalign{\smallskip}
          Method & FID \\
          \noalign{\smallskip}
          \hline
          \noalign{\smallskip}
          StarGAN~\cite{choi2018stargan} & 29.7 \\
          MUNIT~\cite{huang2018multimodal} & 40.2 \\
          StarGANv2*~\cite{choi2019stargan} & 25.6 \\
          Ours & \textbf{14.7} \\
          \hline
          Real images & 3.3 \\
          \hline
        \end{tabular}
      \end{center}
    \end{subfigure}%
    \begin{subfigure}[t]{.5\textwidth}
        \begin{center}
            \caption{Evaluate on Celeba-HQ}
            \label{table:benchmarks-gender-celeba}
            \begin{tabular}{ll}
              \hline\noalign{\smallskip}
              Method & FID \\
              \noalign{\smallskip}
              \hline
              \noalign{\smallskip}
              StarGANv2~\cite{choi2019stargan}\tablefootnote{Official model and weights.} & 27.3 \\
              Ours & \textbf{21.3} \\
              \hline
            \end{tabular}
        \end{center}
    \end{subfigure}%
    \caption{\label{fig:gender_swap} Quantitative comparison with unpaired methods. Unpaired methods trained on the same datasets we evaluate them, although ours trained on FFHQ in both cases. Table~\ref{table:benchmarks-gender-celeba} shows that our method is robust regarding dataset.}
\end{table}

\begin{figure}
    \centering
    \includegraphics[width=\linewidth]{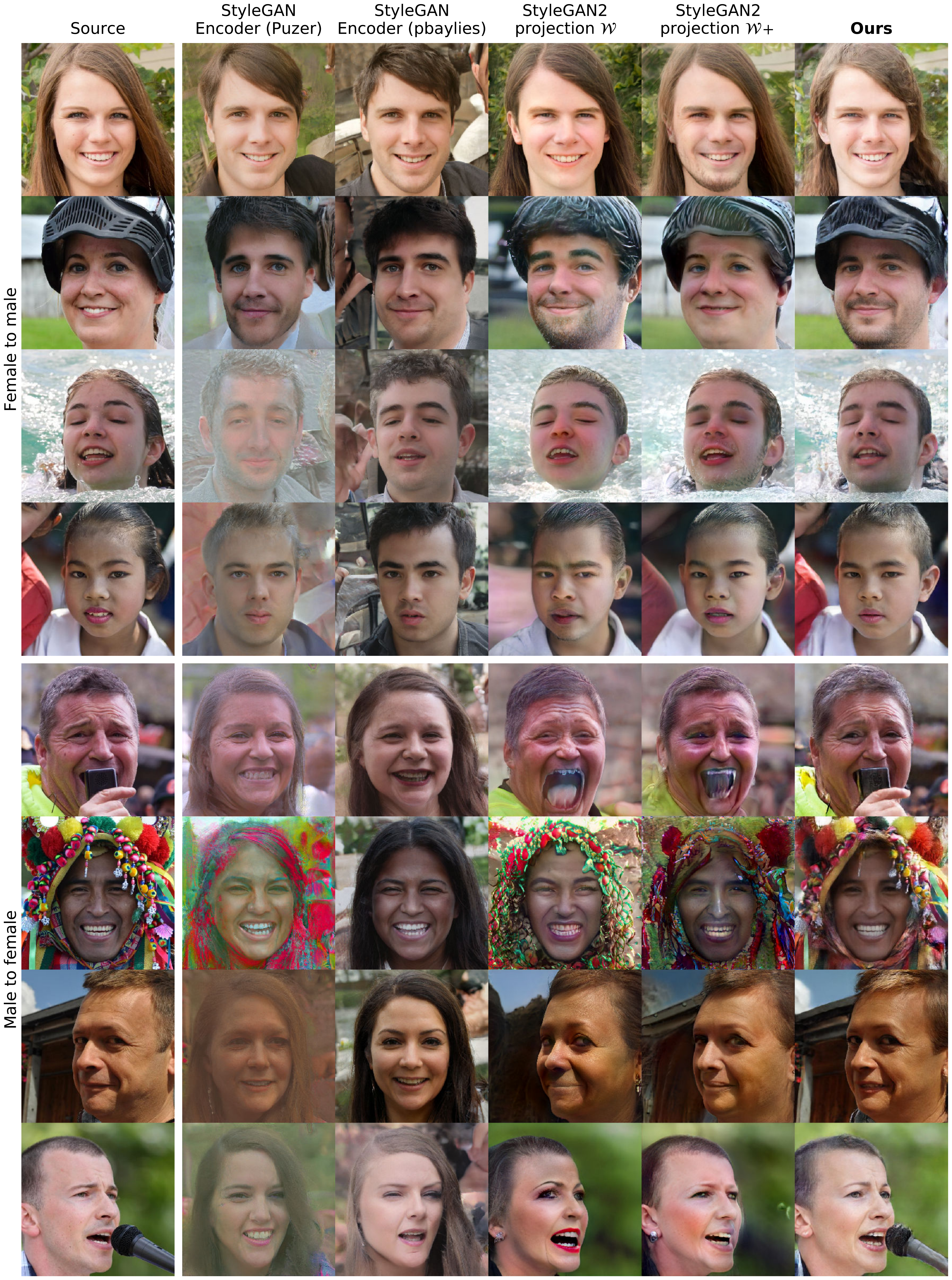}
    \caption{\label{fig:gender_swap_stylegan} Gender transformation: comparison with StyleGAN2 latent code optimization methods. Input samples are real images from FFHQ. Notice that unusual objects are lost with optimization but kept with image-to-image translation}
\end{figure}

Since unpaired methods show significantly worse quality, we put more effort into comparisons between different methods of searching embedding through optimization. We avoid using methods that utilize FID because all of them are based on the same StyleGAN model. Also, FID cannot measure ``quality of transformation'' because it does not check keeping of personality. So we decide to make user study our main measure for all StyleGAN-based methods. Fig.~\ref{fig:gender_swap_stylegan} shows qualitative comparison of all the methods. It is visible that our method performs better in terms of transformation quality. And only StyleGAN Encoder \cite{baylies2019styleganencoder} outperforms our method in realism. However this method generates background unconditionally. 

\begin{table}
    \begin{center}
        \caption{User study of StyleGAN-based approaches. Winrate ``method vs ours''. We measure user study for all StyleGAN-based approaches because we consider human evaluation more reliable measure for perception}
        \label{table:userstudy-gender}
        \begin{tabular}{lll}
          \hline\noalign{\smallskip}
          Method & Quality & Realism \\
          \noalign{\smallskip}
          \hline
          \noalign{\smallskip}
          StyleGAN Encoder (Nikitko) & 18\% & 14\% \\
          StyleGAN Encoder (Baylies) & 30\% & 68\% \\
          StyleGAN2 projection ($\mathcal{W}$) & 22\% & 22\% \\
          StyleGAN2 projection ($\mathcal{W}+$) & 11\% & 14\% \\
          \hline
          Real images & - & 85\% \\
          \hline
        \end{tabular}
    \end{center}
\end{table}

We find that pix2pixHD keeps more details on transformed images than all the encoders. We suppose that this is achieved due to the ability of pix2pixHD to pass part of the unchanged content through the network. Pix2pixHD solves an easier task compared to encoders which are forced to encode all the information about the image in one vector.

Fig.~\ref{fig:gender_swap_i2i} and \ref{fig:gender_swap_stylegan} also show drawbacks of our approach. Vector of ``gender'' is not perfectly disentangled due to some bias in attribute distribution of FFHQ and, consequently, latent space correlation of StyleGAN\cite{shen2020interpreting}. For example, it can be seen that translation into female faces can also add smile.

We also encounter problems of pix2pixHD architecture: repeated patterns, light blobs and difficulties with finetuning 1024x1024 resolution. We show an uncurated list of generated images in supplementary materials.

\subsubsection{Aging/rejuvenation} To show that our approach can be applied for another image-to-image transform task, we also carry out similar experiment with face age manipulation. First, we estimate age for all generated images, then group them into several bins. After that, for each bin we find vectors of ``+2 bins'' and ``-2 bins''. Using these vectors, we generate united paired dataset. Each pair contains younger and older versions of the same face. Finally, we train two pix2pixHD networks, one for each of two directions.
\begin{figure}
    \centering
	\includegraphics[width=\linewidth]{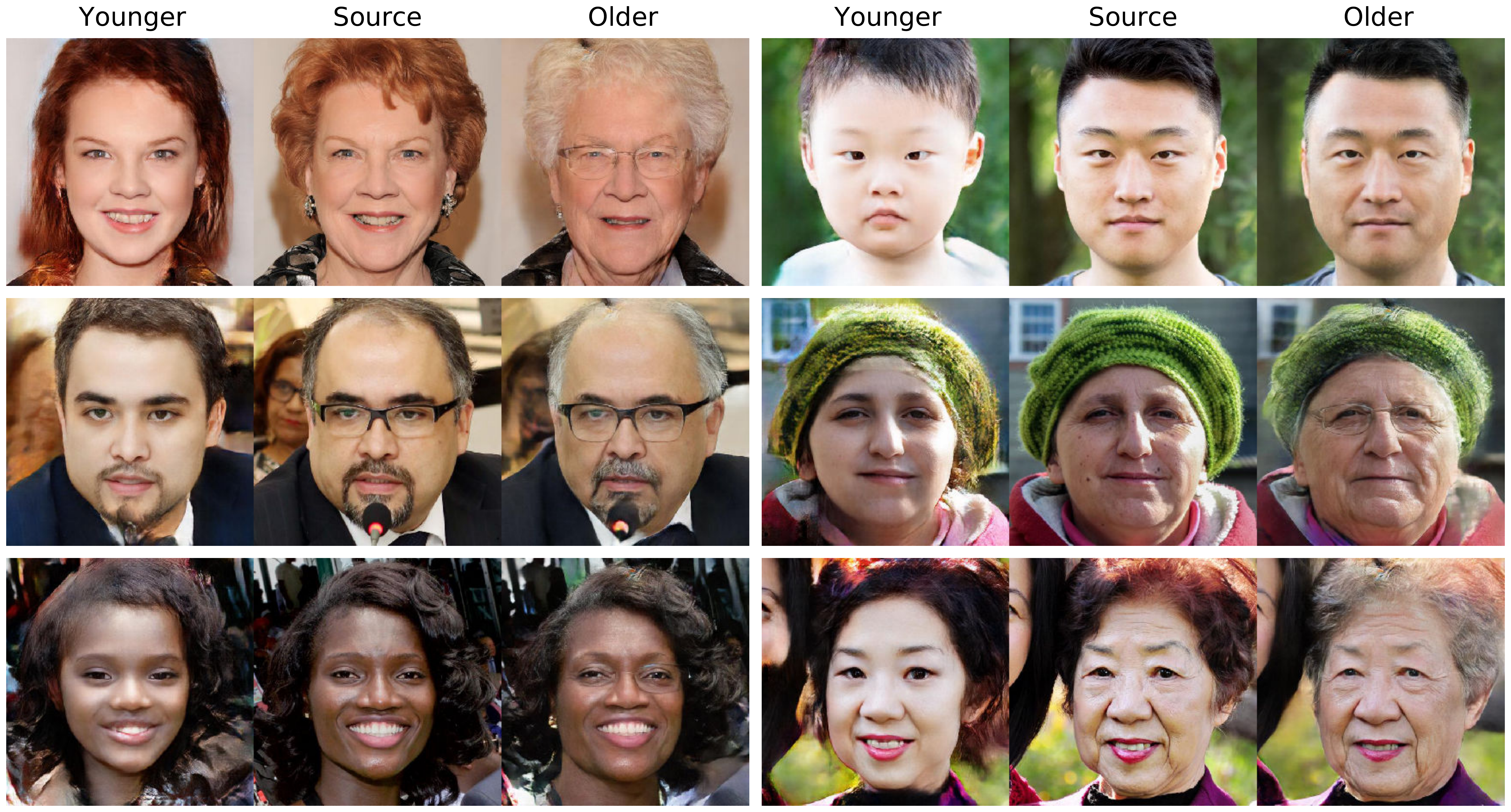}
	\caption{\label{fig:aging} Aging/rejuvenation. Source images are sampled from FFHQ}
\end{figure}
Examples of the application of this approach are presented in Fig. \ref{fig:aging}.

\subsection{Distillation of style mixing}
\subsubsection{Style mixing and face morphing} There are 18 AdaIN inputs in StyleGAN2 architecture. These AdaINs work with different spatial resolutions, and changing different input will change details of different scale. The authors divide them into three groups: coarse styles (for $4^2$ -- $8^2$ spatial resolutions), middle styles ($16^2$ -- $32^2$) and fine styles ($64^2$ -- $1024^2$). The opportunity to change coarse, middle or fine details is a unique feature of StyleGAN architectures.

\begin{figure}
    \centering
	\includegraphics[width=.96\linewidth]{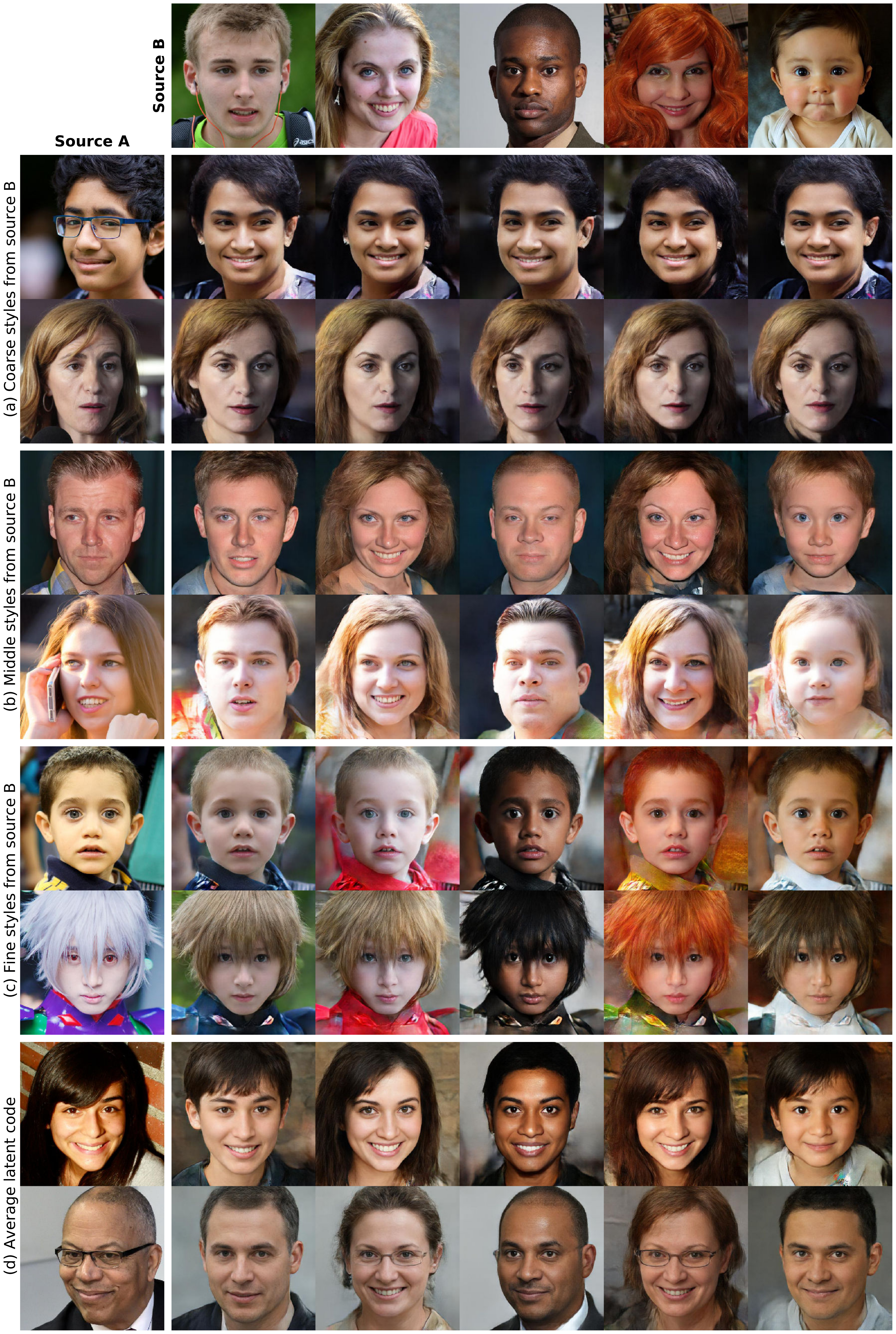}
	\caption{\label{fig:style_mixing} Style mixing with pix2pixHD. (a), (b), (c) show results of distillated crossover of two latent codes in $\mathcal{W}+$, (d) shows result of average latent code transformaiton. Source images are sampled from FFHQ}
\end{figure}

We collect datasets of triplets (two source images and their mixture) and train our models for each transformation. We concatenate two images into 6 channels to feed our pix2pixHD model. Fig.~\ref{fig:style_mixing}(a,b,c) show the results of style mixing.

Another simple linear operation is to average two latent codes. It corresponds to morphing operation on images. We collect another dataset with triplet latent codes: two random codes and an average one. The examples of face morphing are shown in Fig.~\ref{fig:style_mixing}(d).

\section{Conclusions}
In this paper, we unite unconditional image generation and paired image-to-image GANs to distill a particular image manipulation in latent code of StyleGAN2 into single image-to-image translation. The resulting technique shows both fast inference and impressive quality. It outperforms existing unpaired image-to-image models in FID score and StyleGAN Encoder approaches both in user study and time of inference on gender swap task. We show that the approach is also applicable for other image manipulations, such as aging/rejuvenation and style transfer.

Our framework has several limitations. StyleGAN2 latent space is not perfectly disentangled, so the transformations made by our network are not perfectly pure. Despite the latent space is not disentangled enough to make pure transformations, impurities are not so severe. 

We use only pix2pixHD network although different architectures fit better for different tasks. Besides, we distil every transformation to a separate model, although some universal model could be trained. This opportunity should be investigated in future studies.

\bibliographystyle{splncs04}
\bibliography{egbib}
\end{document}